\documentclass[letterpaper]{article} 
\usepackage{aaai2026}  
\usepackage{times}  
\usepackage{helvet}  
\usepackage{courier}  
\usepackage[hyphens]{url}  
\usepackage{graphicx} 
\usepackage{natbib}  
\usepackage{caption} 
\usepackage{subcaption}
\usepackage{algorithm}
\usepackage{algorithmic}
\usepackage{amsmath}

\usepackage{newfloat}
\usepackage{listings}
\DeclareCaptionStyle{ruled}{labelfont=normalfont,labelsep=colon,strut=off} 
\floatstyle{ruled}
\newfloat{listing}{tb}{lst}{}
\floatname{listing}{Listing}
\usepackage{booktabs}
\usepackage{xcolor,colortbl}
\usepackage{csquotes}

\newcommand{\answerYes}[1]{\textcolor{blue}{#1}} 
\newcommand{\answerNo}[1]{\textcolor{teal}{#1}} 
\newcommand{\answerNA}[1]{\textcolor{gray}{#1}}

\title{Triangulating Temporal Dynamics in Multilingual Swiss Online News}
\author {
    Victor Bros\textsuperscript{\rm 1, \rm 2},
    Evan Dufraisse\textsuperscript{\rm 3},
    Adrian Popescu\textsuperscript{\rm 3},
    Daniel Gatica-Perez\textsuperscript{\rm 1, \rm 2}  
}
\affiliations {
    \textsuperscript{\rm 1}Idiap Research Institute, Switzerland\\
    \textsuperscript{\rm 2}EPFL, Switzerland\\
    \textsuperscript{\rm 3}Université Paris-Saclay, CEA, LIST, France\\
    victor.bros@idiap.ch, adrian.popescu@cea.fr, gatica@idiap.ch
}

\begin{document}

\maketitle

\begin{abstract}
Analyzing news coverage in multilingual societies can offer valuable insights into the dynamics of public discourse and the development of collective narratives, yet comprehensive studies that account for linguistic and cultural diversity within national media ecosystems remain limited, particularly in complex contexts such as Switzerland. This paper studies temporal trends in Swiss digital media across the country's three main linguistic regions, French, German, and Italian, using a triangulated methodology that combines quantitative analyses with qualitative insights. We collected and processed over 1.7 million news articles (January 2019–January 2023), applying lexical metrics (diversity, density, sentence length), named entity recognition and Wikidata-based linking, targeted sentiment analysis, and consensus-based change-point detection. To enable principled cross-language comparisons and to connect to theories of domestication and cultural proximity, we derive domestication profiles together with a proximity salience ratio. Our analysis spans thematic, recurrent, and singular events (e.g. the British Royal Family, Christmas, Brexit, and the Swiss Wolf).
By integrating quantitative data with qualitative interpretation, we provide new insights into the dynamics of Swiss digital media and demonstrate the usefulness of triangulation in media studies. The findings reveal distinct temporal patterns and highlight how linguistic and cultural contexts influence reporting. For example, French outlets emphasize France’s role in Brexit negotiations, while German and Italian media focused on broader European implications, patterns reflected in our domestication and proximity indices. In contrast, coverage of the Swiss Wolf is strongly localized, with each linguistic region highlighting its own cantons and concerns, and with consistently high Swiss anchoring across languages. Our approach offers a framework applicable to other multilingual or culturally diverse media environments, contributing to a deeper understanding of how news is shaped by linguistic and cultural factors.
\end{abstract}

\section{Introduction}

The digital transition has reshaped how news is produced, distributed, and consumed. Audiences, especially younger ones, favor online formats, pushing legacy outlets toward digital platforms and accelerating the transnational circulation of news \cite{APhenomenologyOfNews, YoungPeopleRead, TheDriversOfGlobalNews}. This environment is also more fragmented, with implications for how users encounter and evaluate information \cite{WhatNewsUsersPerceive}. While widespread \enquote{echo chambers} are contested, pockets of strong ideological segregation exist and can affect public discourse \cite{EchoTunnels, PoliticalPolarizationInOnlineNews}.
Temporal analysis is important for making sense of news dynamics: it reveals agenda fluctuations, framing shifts, and the evolution of sentiment and actors over time \cite{kwak_systematic_2020-1}. Advances in NLP and time‑series analysis enable scalable detection of phases and breaks in coverage \cite{ARecurrentNeuralModel}. In our approach, these tools are tied to classic accounts of agenda‑setting and issue‑attention cycles, which anticipate surges and declines around focusing events and institutionalized rhythms \cite{downs_up_1991}. In multilingual contexts, however, divergences in coverage may reflect linguistic and cultural dynamics, domestication, and cultural proximity, rather than clear-cut fragmentation.

We therefore ground the study in four complementary strands. First, the agenda‑setting and issue‑attention cycles model temporal salience shifts \cite{mccombs_evolution_1993}. Second, domestication and cultural proximity predict that foreign events are reframed through national or local lenses with systematic cross‑linguistic variation \cite{clausen_localizing_2004,straubhaar_cultural_2021}. Third, comparative media systems place Switzerland in the Democratic Corporatist model, shaping professional norms and outlet roles \cite{hallin_comparing_2004}. Fourth, research on segmented and transnational public spheres anticipates partially parallel, linguistically bounded spheres that remain loosely coupled \cite{wessler_transnationalization_2008}. These perspectives motivate our temporal modeling and our operationalization of cross‑linguistic domestication and proximity.

Switzerland offers a distinctive testbed. It is a small, federal, officially multilingual state (German, French, Italian, Romansh, a small minority) with news production and consumption organized along federal and language‑regional lines and persistent, though porous, language cleavages (the \enquote{Röstigraben}) \cite{BFS_Language,linder_swiss_2021}. The market has consolidated around a few private groups (TX Group, CHMedia, Ringier) alongside a strong public service media (SRG SSR) mandated to serve all language regions \cite{FOEG_Yearbook_2024,OFCOM_SRGConcession}. Swiss audiences are simultaneously exposed to proximate foreign media spheres (France, Germany, Italy), reflecting cultural proximity \cite{NewmanDNR2024}. These structural features plausibly shape agenda‑setting and domestication across regions, public service media counterbalance market concentration, while cross‑border ties and federalism support partially segmented but loosely coupled public spheres \cite{blum_medienstrukturen_2003}. At the same time, consolidation and staffing cuts have raised concerns about regional diversity and the emergence of \enquote{news deserts} \cite{SwissPublisherTamedia1, SwissPublisher3, news_desert_abernathy, InvestigatingNewsDeserts}.

Despite this rich context and literature, few studies operationalize agenda‑setting, domestication, and cultural proximity at scale across coexisting linguistic regions within one national media system. We address this gap with a triangulated design that integrates quantitative and qualitative lenses: lexical indices, Wikidata‑linked domestication profiles and a proximity salience ratio, targeted sentiment toward entities, and a consensus‑based change‑point procedure. Triangulation enhances interpretability by situating quantitative signals in event timelines and media routines \cite{TriangulationInSocialResearch, TriangulationNews, samenews}.

We examine temporal trends in Swiss digital media by focusing on different event types, \textit{thematic}, \textit{singular}, and \textit{recurrent}, across the three main language regions. We illustrate the approach with two focal cases, Brexit (international, singular) and the Swiss Wolf (domestic, thematic), and provide parallel analyses for Christmas (domestic, recurrent) and the British Royal Family (international, thematic). This leads us to the following research questions:

\noindent\textbf{RQ1}: What temporal patterns emerge in the coverage of different event types in Swiss online media?

\noindent\textbf{RQ2}: How do temporal trends in media coverage differ across linguistic regions in Switzerland?

\noindent\textbf{RQ3}: How can a triangulated methodology, combining quantitative and qualitative analyses, enhance understanding of media dynamics in a multilingual context?

We contribute to the literature by (1) developing a triangulated approach that integrates quantitative data analysis and qualitative interpretation, including standardized lexical indices, entity-linked domestication and proximity measures, and consensus change-point detection; (2) analyzing media coverage of various event types across French, German, and Italian online news outlets in Switzerland over a four-year period; (3) providing insights into how linguistic and cultural contexts influence news reporting in a multilingual country; and (4) demonstrating the utility of triangulated approaches in media studies, particularly within complex, multilingual environments. Although centered on Switzerland, the framework is portable to other multilingual or culturally segmented media systems (e.g. Belgium, Canada, Spain, and India) where coexisting language spheres shape coverage.

\section{Related Work}
\label{sec:related}

Research on digital news media encompasses media framing effects, event representation, narrative evolution, and computational techniques to analyze news content. In the context of our study on temporal analysis of multilingual news articles, we review literature on media framing and temporal analysis, event detection using NLP methods, multilingual media studies, definitions of media events, triangulation methodologies, and the Swiss media landscape.

\subsection{Media Framing, Agenda-Setting, and Temporal Dynamics in News Studies}

Media framing significantly influences public perception by emphasizing certain aspects of events. 
\citet{kwak_systematic_2020-1} conducted a frame analysis of 1.5 million \emph{New York Times} articles using a media frame classifier, uncovering both short-term fluctuations corresponding to major events and long-term trends, such as the rise of the \enquote{Cultural identity} frame. Similarly, \citet{TopicModelingUncovers} applied topic modeling and change-point analysis to examine shifts in framing of the German Renewable Energy Act over 17 years, highlighting transitions in media coverage from optimism to cost concerns.
In exploring sensitive social issues, \citet{NewsMediaAnd} analyzed media framing of violence against women, finding that national, right-leaning, and conspiratorial news sources employed more stigmatized framings. This underscores the impact of media attributes on representation and influence.
Whereas framing captures how issues are interpreted through specific schemata, agenda-setting concerns what issues receive attention and when, the two traditions are complementary for studying temporal patterns of news salience.
Beyond framing, agenda-setting research provides a foundational account of how media coverage structures issue salience and attention over time \citep{mccombs_evolution_1993}. Downs’ \enquote{issue-attention cycles} anticipate surges in attention around focusing events or institutionalized rhythms, followed by declines even when problems persist \citep{downs_up_1991}. These regularities justify modeling attention as time series and identifying structural breaks: in our analyses, peaks and shifts in article volume, lexical profiles, and entity prominence are read as empirical traces of agenda-building and attention cycling.

Prior research has examined how news coverage unfolds over time and how different issues are reported. For instance, \citet{FindingNarrativesIn} investigated temporal dimensions such as lifespan and burstiness, revealing distinct behaviors across different stories. Additionally, \citet{AssociationAndTemporality} examined Portuguese news articles and related tweets and observed that news publications intensify around key events, while social media discussions are more evenly distributed.
Advancements in natural language processing have enabled automated event detection and analysis in news media. \citet{BetweenArticleAnd} proposed methods to identify news events as an intermediate level between articles and topics, facilitating meaningful comparisons of media coverage and audience exposure. Enhancements to topic modeling for short texts were introduced by \citet{EnhancingTopic}, who integrated human judgment to improve similarity between text pairs, thereby improving accuracy on sources such as Twitter.
Beyond event detection and topic modeling, other computational approaches have focused on understanding the evolution and propagation of sentiment and information in news coverage.
For example, \citet{LostInPropagation} found that sentiment diminishes in news coverage compared to source press releases but increases in social media interactions. Their work highlights complexities in how content evolves across different platforms.

These methodologies inform our approach, which employs NLP techniques, such as topic modeling and sentiment analysis, combined with change-point detection to automatically identify trends and phases in news coverage.

\subsection{Multilingual Media Studies \& Swiss Context}

Comparative media systems research situates Switzerland in the Democratic Corporatist model, characterized by strong public service media, professionalized journalism, and historically robust press traditions \citep{hallin_comparing_2004}. For Switzerland specifically, scholarship documents distinctive media structures and regulatory traditions \citep{blum_medienstrukturen_2003} and shows that media type often explains variance in \enquote{hard news} orientation more strongly than ownership or language region \citep{udris_mapping_2020}. This evidence tempers purely linguistic explanations by foregrounding system-level attributes, e.g. public service versus commercial outlets, as consequential drivers of coverage. 
At the same time, domestication research argues that foreign events are routinely recontextualized through national or local frames \citep{clausen_localizing_2004}, amid debates over convergence toward a \enquote{global newsroom} versus persistent national particularism \citep{gurevitch_global_1993, hafez_international_1999}. Recent work formalizes \enquote{degrees of domestication,} showing systematic cross-national variation in how foreign events are related to domestic actors, institutions, and everyday life \citep{van_dooremalen_foreignize_2025}. Cultural proximity predicts that outlets emphasize familiar linguistic and cultural referents \citep{straubhaar_cultural_2021}. In multilingual polities, these processes unfold within partially parallel, linguistically segmented public spheres that are nevertheless loosely coupled through translation and cross-lingual gatekeeping \citep{wessler_transnationalization_2008}. Our cross-language comparisons of salience (frequency), referential focus (entities and locations), and sentiment thus probe domestication and proximity within a multilingual media system.

Complementing these system-level considerations, the multilingual nature of global news media presents both challenges and opportunities. \citet{ArmedConflictsIn} utilized unsupervised methods to compile a multilingual corpus on armed conflicts, revealing disparities in coverage even when controlling for violence levels. \citet{BridgingNationsQuantifying} highlighted the role of multilingual users in facilitating cross-lingual information exchange on social media, influencing the diffusion of regional politics and social movements.
In media organizations operating in multilingual contexts, \citet{HowDoesANational} examined Belgium's national news agency and found a heavy reliance on local information subsidies, raising concerns about content diversity. To scale up framing analysis, \citet{AStudyOnScaling} explored the creation of datasets via crowdsourcing across 12 languages, demonstrating the feasibility of multilingual media frame analysis.

Focusing on Switzerland, \citet{udris_mapping_2020} analyzed factors influencing the coverage of major current events and public affairs, finding that media types and ownership explain differences more than language regions. \citet{TransregionalNewsMedia} studied transregional coverage and noted variations by topic and media type. Additionally, \citet{InvestigatingNewsDeserts} applied geoparsing to analyze local news articles, observing signs of news desertification but stable place name diversity.
These studies emphasize the importance of considering linguistic and cultural factors in media analysis, particularly in multilingual contexts such as Switzerland~\citep{bros_decoding_2025}. The Swiss media landscape offers a unique context for examining media quality and coverage diversity. \citet{TheMediatedEngagement} compared Swiss and Chinese media coverage of Switzerland's engagement with the Belt and Road Initiative, revealing differences shaped by ideological positions and cultural contexts.

These insights into the Swiss media landscape underscore the importance of considering multilingual and regional dynamics in media analysis, a central focus of our study.

\subsection{Triangulation Methodologies in Media Studies}

Triangulation refers to the use of multiple methods or sources in research to gain a comprehensive perspective on a phenomenon. In this sense, integrating quantitative and qualitative methods deepens understanding in media research. \citet{TriangulationInSocialResearch} discussed the benefits of triangulation in enhancing interpretation rather than merely validation. In news translation, \citet{TranslationWithout} suggested combining comparative text analysis with fieldwork to contextualize findings. Similarly, \citet{TheConstructionOf} employed a multidisciplinary approach to analyze media coverage of international events, combining discourse analysis with attitudinal studies.
\citet{TriangulationAndIntegration} elaborated on the processes of triangulation, emphasizing data integration across stages to support thoughtful analysis. This methodological approach aligns with our intention to triangulate quantitative automatic analyses with qualitative case studies to interpret media coverage patterns.
Despite rich traditions in agenda‑setting, domestication/cultural proximity, and comparative media systems, few studies operationalize these frameworks at scale across coexisting language regions within a single national media system. Our triangulated design directly addresses this gap.

\section{Methodology}
\label{methodology}

\begin{figure*}[h]
    \centering
    \includegraphics[width=0.99\textwidth]{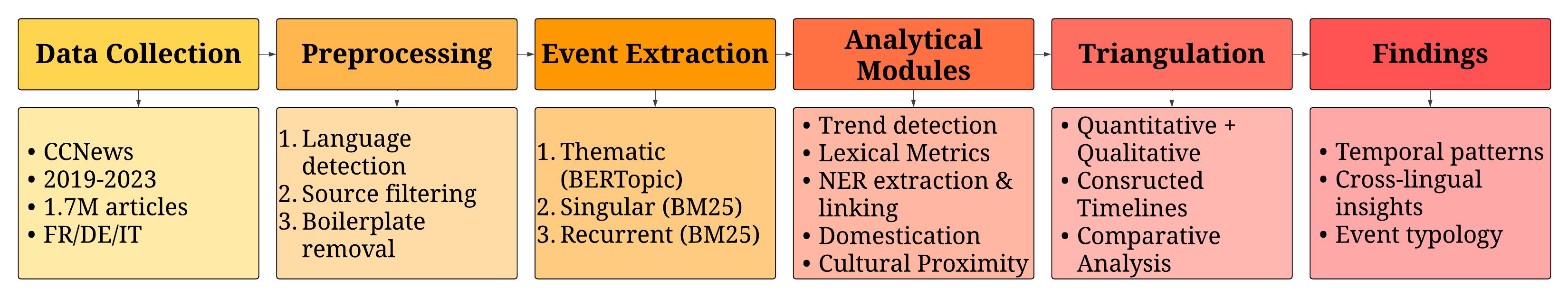}
    \caption{\textbf{Triangulated methodology for temporal analysis of Swiss digital news events.} The diagram summarizes the integrated workflow, combining multilingual data collection, automatic event extraction, quantitative analyses, qualitative interpretation, and triangulated findings on temporal and cross-lingual patterns.}
    \label{fig:method-overview}
\end{figure*}

In this study, we analyze temporal trends in Swiss digital media across the French, German, and Italian languages. We collected a comprehensive dataset of news articles and employed a combination of automatic extraction methods and linguistic analyses to identify and characterize different types of media events, including measurement of lexical diversity and density, named entity recognition and linking with country anchoring, targeted sentiment analysis, and robust, consensus-based change-point detection. To enable principled cross-language comparisons, we standardize lexical metrics within language and operationalize domestication and cultural proximity through two indices derived from Wikidata-linked entities.
We conclude with a triangulation approach that integrates quantitative and qualitative analyses to provide a nuanced understanding of media coverage. Figure~\ref{fig:method-overview} provides an overview of the end‑to‑end pipeline. The remainder of this section details each stage.

\subsection{Data Collection and Processing}

We compiled a dataset of digital news articles from CCNews~\cite{CCNewsHamborg}, a subset of the CommonCrawl repository containing exclusively media outlet content. Articles were collected from January 2019 to January 2023, encompassing a range of Swiss news sources in French, German, and Italian.
Language inference was performed on each article using language detection algorithms to accurately categorize the content~\cite{Fasttext}. Despite initial consideration, we excluded Romansh from our analysis due to the limited number of articles (fewer than 8,000 over four years from a single reliable source), focusing instead on the other Swiss official languages with sufficient data volume.
We selected sources based on three criteria: (i) temporal consistency—sources with a consistent distribution of articles over the entire timeframe to ensure comprehensive coverage; (ii) content scope—generalist news outlets to capture a broad spectrum of topics, excluding niche publications that may not reflect general media trends; and (iii) data quality—exclusion of sources with a high ratio of missing or faulty content ($\geq5\%$). We also screened articles for boilerplate (e.g. repeated footers or newsletter invitations) and removed identifiable boilerplate to enhance data cleanliness.
The corpus contains 1,702,671 articles from 29 unique outlets: 1,247,048 German (18 outlets), 332,822 French (8), and 122,801 Italian (4).

    

\subsection{Definition of Media Events}

Previous studies have provided foundational definitions of media events, notably the work of~\citet{MediaBroadcasting}, who conceptualized media events as exceptional occurrences that interrupt regular programming and command the collective attention of audiences. \citet{RethinkingConceptMediaEvents} expanded on this by introducing a master category of exceptional mediated events, encompassing four main types: media events, media disasters, news events, and pseudo-events. This categorization emphasizes the need for analytical tools that focus on temporality, organization, scale, liveness, and genre in understanding media events within a complex landscape.

While these frameworks provide valuable insights, our study adopts a different approach, focusing specifically on the temporal dimension to accommodate automatic extraction. Using unsupervised techniques, we identify events within our dataset and define media events based on their temporal characteristics and manifestations in digital news articles, categorizing them into three distinct types:

\begin{itemize}
    \item \textbf{Thematic Events}: Events characterized by an overarching theme or topic that evolves over time. These events are connected through common subject matter rather than a singular occurrence and are identified using topic modeling methods, specifically BERTopic~\cite{Bertopic}. Because of their thematic nature, they are more challenging to delineate within clear temporal boundaries.

    \item \textbf{Singular Events}: Unique occurrences within a specific time frame that attract intense media coverage. These events are significant happenings that do not recur frequently, such as major political decisions or significant policy announcements. We extract these events using the BM25 algorithm~\cite{BM25}, employing predefined keywords as queries to retrieve relevant articles. With this approach, we prioritize precision over recall to ensure that the selected articles are highly relevant to the event.

    \item \textbf{Recurrent Events}: Events that occur regularly, often annually, and are anticipated by both media and the public. They include cultural festivals, holidays, and annual conferences. Similar to singular events, we extract recurrent events using the BM25 algorithm with predefined keywords, focusing on events that are known to recur within specific time intervals.
\end{itemize}

Our definitions and categorizations are thus closely aligned with the automatic extraction methods employed. The use of BERTopic for thematic events allows us to uncover latent topics in the dataset without prior knowledge, while BM25 enables us to target specific events based on known keywords. This methodology may result in some overlap between event types, reflecting the complexities of media reporting where an event can have both singular occurrences and thematic elements.
We distinguish targeted retrieval (BM25) and unsupervised discovery (BERTopic) with method‑specific inclusion rules. BM25 events had to appear in all three language regions and meet minimum per‑language volume ($\approx100$ articles for the smallest language, Section~\ref{sec:results}). Queries prioritized precision and were validated by manual spot checks (full lists in Appendix~\ref{app:bm25}). For BERTopic we retained interpretable, adequately sized clusters that surfaced in $\geq2$ linguistic regions with visible temporal structure (Appendix~\ref{app:bertopic_selection}).

\subsection{Lexical Diversity and Density}

We summarize textual variation with three standard metrics used across French, German, and Italian: (i) Type–Token Ratio (TTR) as an indicator of vocabulary variety; (ii) lexical density (share of content words among all tokens); and (iii) mean sentence length (MSL) as a proxy for syntactic complexity \cite{ttr, msl}. Because absolute levels are not directly comparable across languages, we standardize each metric within a language by converting article-level values to $z$-scores relative to that language’s corpus.

To obtain a single summary indicator, we compute a composite standardized index as the equal-weight average of the three language-specific $z$-scores. We aggregate monthly and apply a three-month rolling mean for visualization. All cross-language comparisons rely on standardized scores, a relative-change robustness is developed in the  Appendix~\ref{app:lexical_relchange}.

\subsection{Named Entity Recognition and Linking}

We extract named entities with GLiNER \cite{Gliner}, focusing on three canonical categories widely used in media analysis and an additional custom one, enabled by the generalist model used: \textit{Person}, \textit{Location}, \textit{Organization}, and \textit{Event}. In line with news usage, categories are functional (e.g. a non-human actor may be treated as a Person if framed as an agent in coverage. In the Results, \enquote{Wolf} may be classified as a \textit{Person} when it is treated as an actor in the narrative of the news articles.).

To consolidate variant surface forms and enable cross-language comparisons, we link extracted mentions to Wikidata IDs using mGenre \cite{MGenre} and retain links with confidence $\geq 0.5$. Linked entities are aggregated at weekly or monthly levels and serve as inputs to domestication and cultural proximity indicators. Model benchmarks, threshold checks, and additional notes on cross-language robustness are provided in Appendices~\ref{app:gliner_bench} and~\ref{app:linking_robustness}.

\subsection{Operationalizing Domestication and Cultural Proximity}

To connect our quantitative analysis to domestication and cultural proximity (see Section~\ref{sec:related}), we derive two complementary measures from linked entities: a domestication profile and a proximity salience ratio. Both are computed at the article level and then aggregated weekly.

\paragraph{Domestication profile.} For each article, we compute the proportions of linked mentions anchored to (i) Switzerland, (ii) the language’s proximate neighbor(s) (France for FR, Germany/Austria for DE, Italy for IT), and (iii) other foreign actors (including EU institutions and the UK). Country anchors are inferred from Wikidata properties associated with the linked items. Aggregation yields weekly shares by language. We minimally augment recall with short language-specific cue lists, details are in Appendix~\ref{app:linking_robustness}.

\paragraph{Cultural proximity ratio (PSR).} We quantify the relative salience of proximate versus other foreign references using a smoothed log ratio, as $\mathrm{LPSR}=\log\!\big((m_{\text{prox}}+1)/(m_{\text{other}}+1)\big)$
where $m_{\text{prox}}$ counts mentions anchored to the language’s proximate partner(s) and $m_{\text{other}}$ counts foreign mentions that are neither proximate nor Swiss. We compute weekly averages by language.

\subsection{Sentiment Analysis}

To assess sentiment toward entities, we follow recent guidelines for reliable sentiment measurement in news contexts~\cite{hamborg-donnay-2021-newsmtsc, bestvater_monroe_2023} and employ targeted sentiment classification (TSC) models. Specifically, we utilize the training dataset and protocol from \citet{dufraisse-etal-2023-mad} to train encoder-based models for French, German, and Italian. Sentiment inference is performed as a 3-class task (positive, neutral, negative) for each entity mention. To compute an aggregated sentiment score for each entity, we subtract the negative‑class probability from the positive‑class probability (positive values indicate positive sentiment), yielding a continuous score that reflects the overall sentiment polarity associated with the entity across a sample of occurrences.

\subsection{Change Point Detection}
\label{subsec:cpd}

To identify meaningful shifts in coverage that relate to agenda-building and attention cycles, we adopt a robust, consensus-based change-point detection (CPD) procedure. We build weekly time series for article frequency, but can be generalized to various time series (e.g. Lexical Metrics, Named Entity Frequency). For display, we show weekly counts together with a three-week moving average and a LOESS trend. CP detection is run on the smoothed series (three-week moving average). We generate candidate change points using multiple detectors (PELT with an RBF cost across a small penalty grid, Binary Segmentation with the number of breakpoints matched to PELT’s baseline, and Bayesian Online CPD with a constant hazard), and then cluster all candidate cut points within $\pm 1$ week. We retain the cluster median as a \emph{consensus} change point if it is supported by at least two detectors.

\subsection{Triangulation Approach}

Our study employs a triangulation methodology that integrates quantitative analyses with qualitative insights to provide a comprehensive understanding of media event coverage, while avoiding reliance on any single indicator. We proceed in five coordinated steps: (1) \textbf{Quantitative analysis}: track article frequencies, compute lexical metrics, extract named entities and link them to derive domestication and cultural proximity profiles, estimate targeted sentiment toward entities, and apply change-point detection. (2) \textbf{Contextualization with a priori knowledge}: compile concise event timelines (key dates, milestones, institutional rhythms, background) to anchor statistical variation. (3) \textbf{Qualitative interpretation}: align shifts, peaks, and breakpoints with timelines and media routines, assessing plausible mechanisms (policy windows, seasonality, focusing events). (4) \textbf{Comparative analysis}: systematically compare across languages and event types, with particular attention to domestication profiles and LPSR to probe cultural proximity, alongside contrasts in salience, referential focus, and sentiment. (5) \textbf{Integration of findings}: synthesize these strands into integrated narratives that characterize phases of coverage and cross-lingual dynamics.

This triangulated design enhances interpretability by situating automatic signals within event chronologies and media practices and by seeking converging evidence across heterogeneous indicators, yielding concise, comparable accounts of temporal structure and domestication in a multilingual news system.

\section{Results}
\label{sec:results}

We analyze media coverage of four events: Brexit, Christmas, the British Royal Family, and the Swiss Wolf. The selection balances event types and extraction methods (singular/recurrent via BM25, thematic via BERTopic). We present full, parallel analyses for two focal events in the main text, Brexit (international, BM25-targeted singular) and the Swiss Wolf (domestic, BERTopic/thematic), and include complete dossiers for the other events in the Appendix~\ref{app:dossier_christmas}~and~\ref{app:dossier_brf}. This structure demonstrates breadth (multi-event coverage) and depth (two detailed cases).
The four event subsets contain 2,093/2,699/801 Brexit articles, 4,226/5,179/2,284 Christmas articles, 764/2,579/0 Royal Family articles, and 2,374/2,918/652 Swiss Wolf articles in French/German/Italian, respectively. The Royal Family topic did not surface in the Italian topic model.


\subsection{Focal Case Studies: Brexit and the Swiss Wolf}

We report parallel results for both focal events: frequency and change points, standardized lexical metrics, domestication profiles, and proximity ratios, named entities with targeted sentiment, and a triangulated narrative. Plotting conventions and CPD visualization details are summarized in Appendix~\ref{app:plots}.

\subsubsection{Article Frequency and Change Points}
\label{sec:freq_cpd_focal}

\begin{figure}[h]
    \centering
    \includegraphics[width=0.435\textwidth]{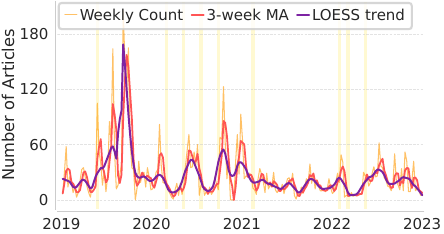}
    \caption{Weekly \textit{Brexit} article counts. Shaded bands indicate consensus change points.}
    \label{fig:brexit_article_distribution}
\end{figure}

\begin{figure}[h]
    \centering
    \includegraphics[width=0.435\textwidth]{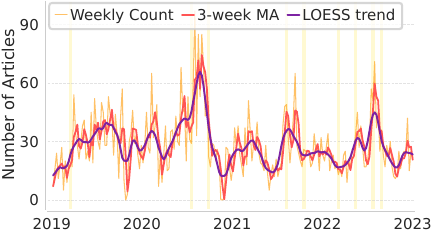}
    \caption{Weekly \textit{Swiss Wolf} article counts. Shaded bands indicate consensus change points.}
    \label{fig:wolf_article_distribution}
\end{figure}

Weekly article counts with consensus change-point detection delineate phases of coverage (see Change Point Detection~\ref{subsec:cpd}). Shaded bands in Figures~\ref{fig:brexit_article_distribution} and~\ref{fig:wolf_article_distribution} mark consensus change points. Implementation choices and plotting settings are in Appendix~\ref{app:plots}.

\paragraph{\textit{Brexit (singular, BM25-targeted).}} 

Figure~\ref{fig:brexit_article_distribution} shows a classic issue‑attention cycle with a small build‑up in early 2019, followed by a sharp regime shift in mid‑2019 around the UK leadership change. The most intense phase spans autumn 2019 to early 2020, with multiple consensus change points bracketing the prorogation crisis, the December 2019 general election, and the UK’s formal withdrawal from the EU (31 January 2020). Subsequent segments capture the negotiation and implementation period: breakpoints in late 2020 align with the end‑of‑year deadline and signature of the Trade and Cooperation Agreement, while early 2021 marks the start of the post‑Brexit regime and related frictions. After a lower and more stable plateau through 2021–early 2022, a final cluster of change points in mid‑to‑late 2022 coincides with the UK leadership turmoil (Johnson’s resignation, Truss’s short premiership, Sunak’s appointment) and a tapering of attention thereafter. The LOESS trend mirrors these phases, with the global maximum occurring around the late‑2019/early‑2020 interval.

\paragraph{\textit{Swiss Wolf (thematic, BERTopic-derived).}} 

In Figure~\ref{fig:wolf_article_distribution}, coverage increases steadily from early 2019 and culminates in a pronounced burst in late summer–early autumn 2020, bounded by consensus change points around the national referendum on the hunting law (27 September 2020). A marked break and decline at the turn of 2020/2021 signal the end of the referendum phase. Thereafter, attention becomes cyclical, with recurrent summer–autumn pulses that coincide with the alpine grazing season and reports of livestock depredations. Change‑point clusters in 2021–2022 track regulatory steps and cantonal authorizations for culls, the largest post‑referendum surge in late summer–autumn 2022 corresponds to a series of high‑salience incidents and federal/cantonal decisions, followed by a gradual easing toward the end of the observation window. The smoothed trend thus reflects a domestically anchored, seasonally patterned attention cycle punctuated by national policy episodes.

\subsubsection{Standardized Lexical Complexity Over Time}
\label{sec:lexical_focal}

We summarize lexical variation with a composite standardized index (within‑language $z$‑scored TTR, lexical density, and mean sentence length; higher values indicate more varied/dense/longer language). A robustness check using relative changes yields the same ordering (Appendix~\ref{app:lexical_relchange}).

\begin{figure}[h]
    \centering
    \includegraphics[width=0.415\textwidth]{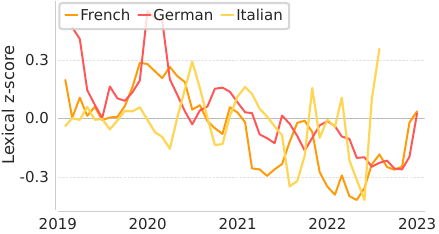}
    \caption{Standardized composite lexical index for \textit{Brexit} (monthly, 3‑month MA).}
    \label{fig:brexit_lexical_metrics}
\end{figure}

\paragraph{\textit{Brexit.}} Figure~\ref{fig:brexit_lexical_metrics} shows elevated lexical complexity in late 2019–early 2020 in French and German outlets, consistent with legalistic and procedural reporting around the election, withdrawal date, and the Trade and Cooperation Agreement. From mid‑2020 onward, both languages exhibit a gradual drift toward simpler writing (indices trending to or below zero), in line with the routinization of coverage as the story moves from negotiation to implementation. A mild rebound in 2022 coincides with the UK leadership crisis, when interpretive pieces and retrospectives temporarily increased linguistic complexity. The Italian series is more volatile, partly reflecting the smaller sample size, but shows the same broad pattern of a dip through 2020–2021 and a short-lived uptick around mid‑2022. Overall amplitudes remain moderate (typically within $\pm 0.5,z$), indicating changes in article mix and style rather than wholesale shifts in register.

\begin{figure}[h]
    \centering
    \includegraphics[width=0.415\textwidth]{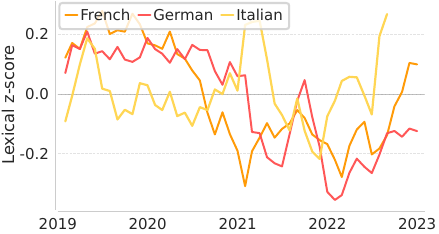}
    \caption{Standardized composite lexical index for \textit{Swiss Wolf} (monthly, 3‑month MA).}
    \label{fig:wolf_lexical_metrics}
\end{figure}

\paragraph{\textit{Swiss Wolf.}} For the \emph{Swiss Wolf} topic (Figure~\ref{fig:wolf_lexical_metrics}), the composite index is comparatively stable and organized in episodes. French- and German‑language coverage sits slightly above baseline through 2019–early 2020 and peaks around the 2020 hunting‑law referendum, when articles include legislative detail and expert commentary. Indices then soften through 2021, a period dominated by brief incident reports and updates, before a partial recovery in late 2022 during renewed regulatory debates and cantonal decisions. The Italian series shows two marked, event‑concordant rises (Q4 2021 and late summer–autumn~2022), consistent with bursts of analysis tied to regional salience. As with Brexit, magnitudes are modest, suggesting that observed fluctuations chiefly reflect shifts in the mix between short news items and longer explanatory pieces rather than systematic language‑specific baselines, which are controlled for by standardization.

\subsubsection{Domestication Profile and Cultural Proximity (LPSR)}
\label{sec:domestication_focal}

\begin{figure}[h]
    \centering
    \includegraphics[width=0.44\textwidth]{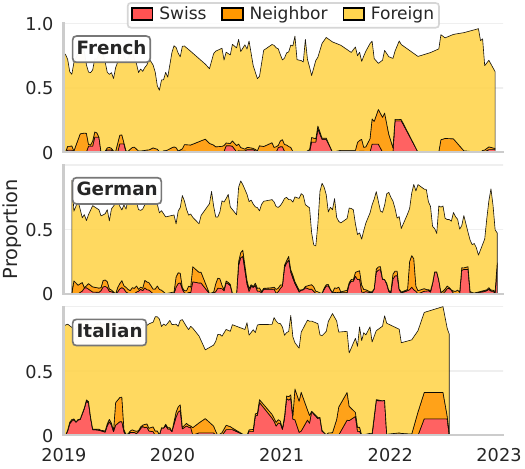}
    \caption{Weekly domestication profile for \textit{Brexit}.}
    \label{fig:brexit_domestication}
\end{figure}

\begin{figure}[h]
    \centering
    \includegraphics[width=0.44\textwidth]{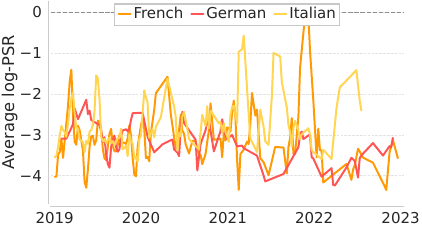}
    \caption{LPSR over time for \textit{Brexit} (weekly averages).}
    \label{fig:brexit_lpsr}
\end{figure}

\begin{figure}[h]
    \centering
    \includegraphics[width=0.45\textwidth]{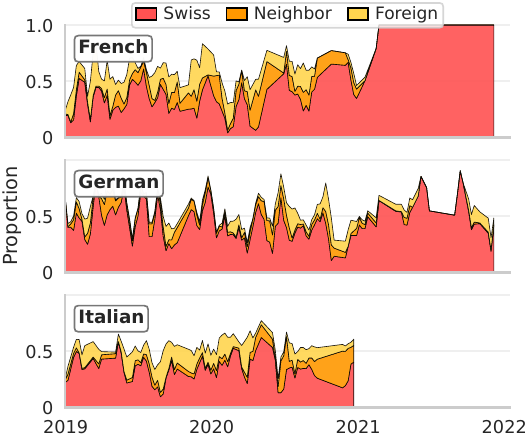}
    \caption{Weekly domestication profile for \textit{Swiss Wolf}.}
    \label{fig:wolf_domestication}
\end{figure}

\begin{figure}[h]
    \centering
    \includegraphics[width=0.44\textwidth]{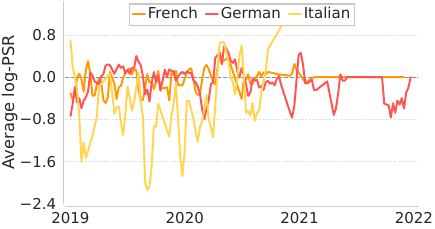}
    \caption{LPSR over time for \textit{Swiss Wolf} (weekly averages).}
    \label{fig:wolf_lpsr}
\end{figure}

We report weekly domestication shares (Swiss, neighbor, foreign‑other) alongside the log Proximity Salience Ratio (LPSR). Figures~\ref{fig:brexit_domestication}–\ref{fig:wolf_lpsr} show small multiples by language.

\paragraph{\textit{Brexit.}}


As expected for an international story, foreign anchoring dominates across languages (large \enquote{foreign} strata in Figure~\ref{fig:brexit_domestication}). French‑language outlets display the most pronounced and sustained \enquote{neighbor} share, particularly during negotiation and decision windows (late~2019; late~2020), whereas German‑ and Italian‑language coverage leans more heavily on EU institutions and the UK (\enquote{foreign‑other}), with only episodic domestication spikes around Swiss impact pieces (e.g. market access, bilateral relations).

The LPSR trajectories in Figure~\ref{fig:brexit_lpsr} are consistently negative, indicating that references to proximate partners are outweighed by other foreign actors. The French and Italian series show brief upward movements toward zero in negotiation or leadership crisis phases, consistent with the heightened salience of proximate elites. The German series, by contrast, remains more persistently negative through 2021–2022, in line with an EU‑level frame rather than a Germany‑specific one. Overall, the pattern aligns with domestication and cultural‑proximity expectations: a predominantly foreign, supranational frame with language‑specific tilts toward proximate actors, strongest in French coverage.

\paragraph{\textit{Swiss Wolf.}}


The \emph{Swiss Wolf} topic is overwhelmingly domesticated in all three languages (Figure~\ref{fig:wolf_domestication}). Swiss anchoring rises markedly around the 2020 hunting‑law referendum and remains elevated thereafter, with French coverage approaching saturation in several stretches and German/Italian coverage maintaining high Swiss shares while still reflecting regional variation. Neighbor references are marginal and largely confined to episodic cross‑border mentions (e.g. Alpine packs, regulatory comparisons), which never displace the Swiss focus.

Consistent with this strong domestication, LPSR values hover near zero or mildly negative in Figure~\ref{fig:wolf_lpsr}, reflecting the rarity of foreign mentions and the absence of a proximate‑neighbor lens. Occasional positive blips, most visible in the Italian series in late~2021–2022, coincide with short bursts of articles referencing Italy or Austrian cases, but these remain temporary and bounded. Taken together, the domestication profile and LPSR confirm that the \emph{Swiss Wolf} is framed as a Swiss, often cantonal, issue with limited reliance on proximate foreign referents.

\subsubsection{Entity Salience and Targeted Sentiment}
\label{sec:ner_focal}

We track mention counts for canonical entities (persons, organizations, locations, event/legislative labels) with Wikidata linking, and targeted sentiment for entity mentions using a 3‑class TSC model, aggregated weekly.

\begin{figure}[h]
    \centering
    \includegraphics[width=0.425\textwidth]{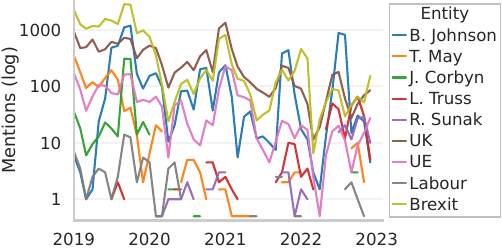}
    \caption{Monthly frequency of \textit{Brexit} key named entities in French-language coverage.}
    \label{fig:brexit_ner_timeline}
\end{figure}

\paragraph{\textit{Brexit.}} Figure~\ref{fig:brexit_ner_timeline} shows that entity salience follows the phases identified by change‑point detection. Mentions of UK political leaders (Theresa May, Boris Johnson, Liz Truss, Rishi Sunak) and EU actors (e.g. the European Commission and Council) surge during the 2019 leadership change, the December 2019 election, the 31 January 2020 withdrawal, and the end‑2020 agreement deadline. Peaks for Brussels and EU institutions align with negotiation windows, whereas later flares in 2022 reflect leadership turnover in the UK. The Italian series is more irregular due to lower volume, but displays the same alignment of entity spikes with milestone weeks.

\begin{figure}[h]
    \centering
    \includegraphics[width=0.45\textwidth]{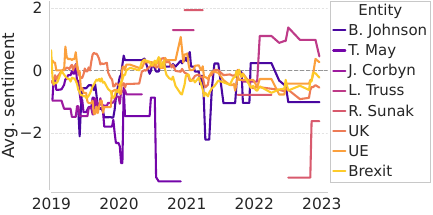}
    \caption{Evolution of weekly sentiment scores for \textit{Brexit}-event entities in French-language coverage.}
    \label{fig:brexit_sentiment}
\end{figure}

Targeted sentiment (Figure~\ref{fig:brexit_sentiment}, French subset for illustration) is slightly negative overall for the event and for most political figures. Theresa May’s trajectory trends downward through mid‑2019, Boris Johnson’s polarity oscillates around mildly negative with dips at moments of procedural crisis, Liz Truss registers a short positive blip at appointment followed by a rapid fall at resignation. Rishi Sunak remains closer to neutral in the observed window. The \enquote{Union européenne} and \enquote{Royaume‑Uni} entities hover near neutral to slightly negative, consistent with conflict‑heavy reporting and economic‑impact frames during peak phases.

\begin{figure}[h]
    \centering
    \includegraphics[width=0.43\textwidth]{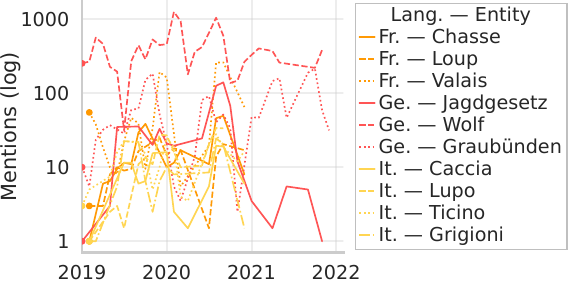}
    \caption{Monthly frequency of \textit{Swiss Wolf} key named entities across all languages.}
    \label{fig:wolf_ner_timeline}
\end{figure}

\paragraph{\textit{Swiss Wolf.}} Figure~\ref{fig:wolf_ner_timeline} plots entity salience on a log scale and highlights the strong domestication of this topic. Mentions of the animal entity (Loup/Wolf/Lupo) and of legislative or hunting labels (Chasse/Jagdgesetz/Caccia) rise sharply around the 2020 hunting‑law referendum, then return in seasonal waves that coincide with the alpine grazing period. Location entities are distinctly language‑regional: \emph{Valais} in French, \emph{Graub\"unden} in German, \emph{Ticino/Grigioni} in Italian, underscoring cantonal anchoring and local incident reporting.

\begin{figure}[h]
    \centering
    \includegraphics[width=0.45\textwidth]{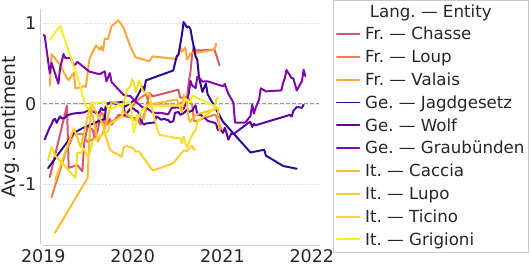}
    \caption{Evolution of weekly sentiment scores for key entities in \textit{Swiss Wolf} coverage (by language).}
    \label{fig:wolf_sentiment}
\end{figure}

The multi‑language targeted sentiment series in Figure~\ref{fig:wolf_sentiment} cluster around neutrality with event‑concordant swings. Negative excursions accompany periods dominated by livestock attacks or authorization of culls, while brief positive turns appear during conservation or management announcements. German‑language lines tied to the legislative label (\emph{Jagdgesetz}) show clearer spikes around national decision points, French and Italian series display smaller amplitudes but follow the same referendum and seasonal rhythm. Taken together with the domestication indices, these trajectories indicate that the Swiss Wolf is framed primarily through local impacts and regulatory episodes, with sentiment responding to incident‑driven controversy rather than to a stable ideological tone.

\subsubsection{Triangulated Case Narratives}
\label{sec:triangulated_focal}

Bringing the signals together, \textit{Brexit} displays a textbook issue–attention cycle. Consensus change points bracket leadership change, the December 2019 election, the 31 January 2020 withdrawal, and the end‑2020 agreement, followed by a lower, steadier implementation phase with brief resurgences during the 2022 leadership turmoil. The standardized lexical index is highest during the procedural/negotiation peak and gradually declines as coverage routinizes. Domestication and proximity metrics indicate a predominantly foreign frame with language‑specific tilts: French outlets temporarily raise the neighbor share and LPSR around negotiation milestones, whereas German and Italian coverage remains anchored in EU/UK institutions. Entity timelines align with these phases (leaders and EU bodies peaking at key junctures), and targeted sentiment is mildly negative for most actors, with short‑lived positive inflections at appointments that fade at moments of crisis.

By contrast, the \textit{Swiss Wolf} case is strongly domesticated across the board. Attention peaks around the 2020 hunting‑law referendum and returns seasonally with the alpine grazing period, lexical complexity is comparatively stable with modest upticks during legislative debates. The domestication profile is dominated by Swiss anchoring, and LPSR hovers near zero or negative, reflecting the marginal role of proximate foreign referents. Entity salience is cantonal and language‑regional (Valais, Graub\"unden, Ticino/Grigioni), and targeted sentiment oscillates around neutrality with sharper negative dips during livestock attacks or culling authorizations and small positive pulses around management or conservation announcements.

Together, these parallel narratives show how agenda dynamics interact with domestication: a transnational negotiation covered through supranational actors versus a localized wildlife‑policy issue rooted in cantonal experience.

\subsection{Cross-Event Synthesis}

Across the four cases, we observe distinct temporal \enquote{signatures} by event type. The international singular event (Brexit) shows spikes aligned with institutional milestones and a strong foreign anchoring, the domestic thematic case (Swiss Wolf) exhibits a referendum peak followed by seasonal pulses with cantonal anchoring, the recurrent cultural event (Christmas) displays stable annual peaks with domesticated coverage, and the international celebrity‑thematic case (British Royal Family) is organized in short, high‑visibility bursts tied to ceremonial or media moments.

Domestication and proximity patterns are consistent with these signatures. Brexit and the British Royal Family are dominated by foreign references with negative LPSR, and only modest, episodic neighbor tilts (strongest in French). Swiss Wolf is overwhelmingly domesticated in all languages, with LPSR hovering near zero or negative. Christmas is locally anchored at the peak while retaining a persistent foreign‑other layer (global retail/travel frames), with LPSR mostly negative and only short‑lived proximate lifts.

Lexical variation is generally modest across events, and higher complexity coincides with procedural or legislative windows (e.g. Brexit peak, referendum phase for Swiss Wolf). Entity salience and targeted sentiment align with the temporal structure of each case: commemorations and festivities trend neutral‑to‑positive, while controversy windows (e.g. leadership crises, culling authorizations, celebrity conflicts) are associated with more negative sentiment. Full details are provided in the event dossiers (Appendix).

\section{Discussion}

\subsection{Interpretation of Findings}

The observed differences in media coverage across regions can be interpreted in light of Switzerland's distinct linguistic and cultural contexts. The focus of French-language media on national actors during Brexit, for example, may reflect a cultural orientation toward France's international role and political leadership. In contrast, the emphasis on European institutions and broader EU implications in German- and Italian-language outlets suggests a perspective more attuned to supranational dynamics and their impact on local communities. These tendencies are reflected in our domestication profiles and proximity salience ratios: Brexit coverage is predominantly anchored in foreign and supranational actors, with only episodic neighbor tilts, most visible in French outlets.
The relatively uniform coverage of Christmas across all language regions, with each highlighting local customs and commercial activities, points to the shared cultural significance of the holiday while also reflecting the diversity of regional traditions within Switzerland. At the same time, the domestication profiles show a clear rise in Swiss anchoring around the seasonal peak, with foreign references primarily tied to global cues.

For internationally themed celebrity coverage, such as the British Royal Family, attention is organized in short, high-visibility bursts tied to ceremonial events and media releases. This pattern is accompanied by overwhelmingly foreign anchoring and negative proximity ratios across languages, consistent with an international, entertainment-adjacent framing rather than a neighbor-centric one.
Coverage of the Swiss Wolf event demonstrates how local realities and concerns are foregrounded in each linguistic community, while national debates can unify coverage across regions during key moments. The prominence of cantonal entities and legislative labels, the consistently high Swiss anchoring, and proximity ratios near zero underscore a domesticated, policy-centered framing with seasonal pulses. Targeted sentiment reinforces this picture: it oscillates around neutrality with sharper negative dips during livestock attacks or culling authorizations and small positive pulses around management or conservation announcements.

Taken together, the change-point detection, lexical indices, entity timelines, and targeted sentiment cohere into parallel narratives: a transnational negotiation framed through supranational and UK actors versus a localized wildlife-policy issue rooted in cantonal experience.

\subsection{Influence of Linguistic and Cultural Contexts}

The findings underscore the significant influence of linguistic and cultural contexts on media coverage in Switzerland. Despite operating within a shared national framework, each language community exhibits unique reporting styles and emphases, tailoring content to resonate with audience interests, cultural values, and societal priorities. The domestication profiles and proximity ratios align with expectations from cultural proximity and segmented public sphere research: French outlets display stronger episodic neighbor orientation for Brexit, whereas German and Italian coverage is more persistently anchored in EU-level frames; for the Swiss Wolf, domestication dominates in all three languages with marked cantonal anchoring.
The linguistic features inherent in each language also impact readability and comprehension. The stability of standardized lexical indices suggests consistent outlet language use. Where variation occurs, it aligns with shifts in the article mix (e.g. from brief incident reports to explanatory or legislative pieces) rather than wholesale shifts in register. The decrease in lexical complexity during Brexit’s implementation phase, for instance, may reflect routinization as audiences grow familiar with the topic.
These insights reinforce the value of a triangulated methodology that combines quantitative and qualitative approaches, enabling a more nuanced understanding of media dynamics in multilingual societies and connecting system-level structures to observable content indicators.

\subsection{Limitations of the Study}

While this study provides valuable insights into media coverage patterns, several limitations should be acknowledged. Firstly, the dataset is limited to articles included in CCNews, which does not contain the entirety of media output during the study period. This could result in incomplete coverage of certain events or underrepresentation of specific outlets.
Secondly, the selection of events and keywords for analysis was predetermined for targeted retrieval, which may introduce bias. The reliance on BM25 with predefined keywords favors precision over recall, potentially overlooking relevant articles that discuss events using different terminology. For thematic topics derived from BERTopic, we retained clusters that were interpretable and of adequate size, a choice that may introduce researcher degrees of freedom.
Thirdly, the sentiment analysis, Named Entity Recognition, and entity linking depend on model accuracy. Although we mitigate errors by aggregating results and retaining linked mentions with sufficient confidence, mislinking or missed anchors may affect domestication profiles and proximity ratios. The compact lexica used to detect Swiss and proximate cues provide recall gains but may introduce heuristic bias at the margins.
Additionally, the lexical complexity metrics provide a general overview but may not capture deeper linguistic features such as tone, metaphor, or idiomatic expressions that could influence reader interpretation. Our treatment of discourse is intentionally minimal, focusing on lexical variation, entity anchoring, and sentiment, while richer discourse layers lie outside the present scope. Standardization enables cross-language comparisons but abstracts from absolute language-specific baselines.
Finally, while the comparative analysis offers insights into linguistic and cultural influences, it does not account for other factors such as differences in editorial policies, journalistic standards, media type, or ownership structures. The identification of change points may also vary with smoothing choices and penalty settings, even under a consensus-based procedure. Future studies could address these limitations by expanding the dataset to include a wider range of sources, refining language and linking resources, and incorporating additional outlet-level covariates to deepen the analysis. 

\section{Conclusion}

In this study, we analyzed temporal trends in Swiss digital media across French, German, and Italian language regions by focusing on different types of events. We addressed the research questions as follows:

For \textbf{RQ1}, we showed that distinct temporal patterns are associated with different event types. Recurrent events, such as Christmas, displayed regular annual peaks, reflecting predictable media cycles tied to seasonal or scheduled events. Singular events, such as Brexit, had spikes in coverage aligned with specific institutional milestones, which our consensus change-point procedure captured as structural breaks. Thematic events, such as the British Royal Family and the coexistence with the Swiss Wolf, exhibited coverage peaks corresponding to identifiable phases (e.g. legislative debates and seasonal pulses for the Wolf, ceremonial and media releases for the Royal Family).

Addressing \textbf{RQ2}, we observed differences in media coverage among the French, German, and Italian language regions. These differences were influenced by linguistic and cultural contexts, affecting the emphasis and framing of events. For instance, French-language media tended to focus on France's role in international events, while German- and Italian-language media highlighted broader European perspectives or national implications. Our domestication profiles and proximity ratios corroborate these tilts, distinguishing foreign, supranational anchoring for international stories from strong Swiss anchoring in domestically rooted issues.

Regarding \textbf{RQ3}, the triangulated methodology proved effective in enhancing the understanding of media dynamics. By integrating article frequency, lexical indices, NER and linking, targeted sentiment, and change-point detection with qualitative interpretation, we gained deeper insights into how media narratives evolve over time. This allowed us to contextualize quantitative findings within real-world events and cultural nuances, revealing how linguistic and cultural factors influence reporting in a multilingual society.

In summary, our study demonstrates that temporal patterns in Swiss digital media coverage vary according to event types and linguistic regions. The key takeaways from this research are: (1) linguistic and cultural contexts influence media narratives within Switzerland, shaping the framing and emphasis of events in each language region; (2) combining quantitative and qualitative methods, specifically, lexical indices, domestication and proximity measures, targeted sentiment, and change-point detection, provides a comprehensive framework for analyzing media trends in multilingual environments; and (3) understanding temporal dynamics in media coverage offers valuable insights into how events are reported and perceived over time, informing future research and media practices. While centered on Switzerland, the proposed framework can be adapted to other multilingual environments. Applied to Belgium and Canada, our pipeline simply swaps neighbor sets (FR–NL–DE; US–FR) and incorporates the PSB remit (RTBF/VRT; CBC/Radio‑Canada) as a media‑system factor. 

\section*{Acknowledgments}
This work was supported by the European Union’s Horizon 2020 program through the AI4Media project (No. 951911) and by the EU Horizon Europe program through the ELIAS project (No. 101120237). We thank Alessandro Fornaroli and Enno Hermann for their help in clarifying language-region-specific features of the Swiss media landscape.

\begingroup
\setlength{\bibsep}{0pt plus 0.1ex}
\small
\bibliography{aaai25}
\endgroup

\section*{Paper Checklist}

\begin{enumerate}

\item For most authors...
\begin{enumerate}
    \item Would answering this research question advance science without violating social contracts, such as violating privacy norms, perpetuating unfair profiling, exacerbating the socio-economic divide, or implying disrespect to societies or cultures?\\
    \answerYes{Yes, the study is based exclusively on aggregate, publicly available news data and does not involve personal or sensitive information. See Methodology and Data Collection.}
    
    \item Do your main claims in the abstract and introduction accurately reflect the paper's contributions and scope?\\
    \answerYes{Yes, the claims in the abstract and introduction are consistent with the methods and results.}
    
    \item Do you clarify how the proposed methodological approach is appropriate for the claims made?\\
    \answerYes{Yes, the methodology is described in detail and is appropriate for the research questions. See Methodology and Triangulation Approach.}
    
    \item Do you clarify what are possible artifacts in the data used, given population-specific distributions?\\
    \answerYes{Yes, the paper discusses dataset coverage, possible biases, and limitations. See Limitations of the Study.}
    
    \item Did you describe the limitations of your work?\\
    \answerYes{Yes, limitations are discussed explicitly. See Limitations of the Study.}
    
    \item Did you discuss any potential negative societal impacts of your work?\\
    \answerYes{Yes, the study notes possible biases and underrepresentation, but no direct negative societal impact is identified. See Limitations and Discussion.}
    
    \item Did you discuss any potential misuse of your work?\\
    \answerNo{No, as the work is based on aggregate media analysis and does not present realistic misuses.}
    
    \item Did you describe steps taken to prevent or mitigate potential negative outcomes of the research, such as data and model documentation, data anonymization, responsible release, access control, and the reproducibility of findings?\\
    \answerYes{Yes, only publicly available news data is used, and no personal or sensitive information is processed. See Data Collection and Processing.}
    
    \item Have you read the ethics review guidelines and ensured that your paper conforms to them?\\
    \answerYes{Yes, the study conforms to ethical guidelines as it uses only public, aggregate data.}
\end{enumerate}

\item Additionally, if your study involves hypotheses testing...
\begin{enumerate}
  \item Did you clearly state the assumptions underlying all theoretical results?\\
  \answerNA{NA, as the study is primarily empirical and does not present formal theoretical results.}
  
  \item Have you provided justifications for all theoretical results?\\
  \answerNA{NA}
  
  \item Did you discuss competing hypotheses or theories that might challenge or complement your theoretical results?\\
  \answerYes{Yes, the study is situated within the relevant literature and discusses alternative explanations. See Related Work and Discussion.}
  
  \item Have you considered alternative mechanisms or explanations that might account for the same outcomes observed in your study?\\
  \answerYes{Yes, alternative explanations and limitations are discussed. See Limitations and Discussion.}
  
  \item Did you address potential biases or limitations in your theoretical framework?\\
  \answerYes{Yes, see Limitations of the Study.}
  
  \item Have you related your theoretical results to the existing literature in social science?\\
  \answerYes{Yes, see Related Work and Discussion.}
  
  \item Did you discuss the implications of your theoretical results for policy, practice, or further research in the social science domain?\\
  \answerYes{Yes, implications for future research and media analysis are discussed. See Conclusion and Discussion.}
\end{enumerate}

\item Additionally, if you are including theoretical proofs...
\begin{enumerate}
  \item Did you state the full set of assumptions of all theoretical results?\\
  \answerNA{NA, no formal theoretical proofs are included.}
  
  \item Did you include complete proofs of all theoretical results?\\
  \answerNA{NA}
\end{enumerate}

\item Additionally, if you ran machine learning experiments...
\begin{enumerate}
  \item Did you include the code, data, and instructions needed to reproduce the main experimental results (either in the supplemental material or as a URL)?\\
  \answerNo{No, because the code and data are not being released.}
  
  \item Did you specify all the training details (e.g., data splits, hyperparameters, how they were chosen)?\\
  \answerYes{Yes, relevant details for model training and inference are provided in the Methodology and Sentiment Analysis sections.}
  
  \item Did you report error bars (e.g., with respect to the random seed after running experiments multiple times)?\\
  \answerNo{No, as the main analyses are descriptive and not based on repeated stochastic ML model training.}
  
  \item Did you include the total amount of compute and the type of resources used (e.g., type of GPUs, internal cluster, or cloud provider)?\\
  \answerYes{Yes, see Appendix on computation resources.}
  
  \item Do you justify how the proposed evaluation is sufficient and appropriate to the claims made?\\
  \answerYes{Yes, the evaluation is aligned with the research questions and the triangulated methodology. See Triangulation Approach and Results.}
  
  \item Do you discuss what is \enquote{the cost} of misclassification and fault (in)tolerance?\\
  \answerNo{No, as the main analyses are descriptive and not focused on classification tasks.}
\end{enumerate}

\item Additionally, if you are using existing assets (e.g., code, data, models) or curating/releasing new assets, \textbf{without compromising anonymity}...
\begin{enumerate}
  \item If your work uses existing assets, did you cite the creators?\\
  \answerYes{Yes, all major assets (e.g. CCNews, BERTopic, GLiNER, mGenre) are cited. See Data Collection and References.}
  
  \item Did you mention the license of the assets?\\
  \answerNo{No, the licenses are not specified in the current version.}
  
  \item Did you include any new assets in the supplemental material or as a URL?\\
  \answerNo{No, new datasets or code are not being released.}
  
  \item Did you discuss whether and how consent was obtained from people whose data you're using/curating?\\
  \answerNA{NA, as only public news articles are used.}
  
  \item Did you discuss whether the data you are using/curating contains personally identifiable information or offensive content?\\
  \answerYes{Yes, the dataset contains only public news articles and excludes personal information. See Data Collection and Processing.}
  
  \item If you are curating or releasing new datasets, did you discuss how you intend to make your datasets FAIR (see FORCE11 (2020))?\\
  \answerNA{NA, as no new datasets are being released.}
  
  \item If you are curating or releasing new datasets, did you create a Datasheet for the Dataset (see Gebru et al. (2021))?\\
  \answerNA{NA}
\end{enumerate}

\item Additionally, if you used crowdsourcing or conducted research with human subjects, \textbf{without compromising anonymity}...
\begin{enumerate}
  \item Did you include the full text of instructions given to participants and screenshots?\\
  \answerNA{NA, no human subjects or crowdsourcing involved.}
  
  \item Did you describe any potential participant risks, with mentions of Institutional Review Board (IRB) approvals?\\
  \answerNA{NA}
  
  \item Did you include the estimated hourly wage paid to participants and the total amount spent on participant compensation?\\
  \answerNA{NA}
  
  \item Did you discuss how data is stored, shared, and deidentified?\\
  \answerYes{Yes, only public, non-personal data is used. See Data Collection and Processing.}
\end{enumerate}

\end{enumerate}

\clearpage
\appendix

\setcounter{secnumdepth}{1}
\renewcommand\thesection{\Alph{section}}

\section{List of Selected Sources}

\setlength{\tabcolsep}{1pt}
\begin{table}[h]
    
    \centering
    \rowcolors{2}{}{gray!15}
    {\footnotesize
    \begin{tabular}{lll}
        \toprule
        \textbf{German} & \textbf{French} & \textbf{Italian} \\
        \midrule
        20 Minuten & 20 minutes & Corriere del Ticino \\
        Aargauer Zeitung & 24 heures & Swissinfo \\
        Basler Zeitung & La Gruy\`ere & Ticinolibero \\
        bz Basel & La Tribune de Gen\`eve & TicinOnline \\
        ch.ch & RFJ & \\
        Der Bund & RJB & \\
        FM1Today & RTN & \\
        Finanz und Wirtschaft & Swissinfo & \\
        Grenchner Tagblatt &  & \\
        Hochparterre &  & \\
        Der Landbote &  & \\
        Limmattaler Zeitung &  & \\
        Luzerner Zeitung &  & \\
        Neue Z\"urcher Zeitung &  & \\
        SRF &  & \\
        Swissinfo & & \\
        St. Galler Tagblatt & & \\
        Tages-Anzeiger & & \\
        Z\"urcher Unterländer & & \\
        \bottomrule
    \end{tabular}
    }
    \caption{List of selected news sources by language.}
    \label{tab:sources}
\end{table}

\section{Topic Modeling Parameters}

The topic modeling was performed using the BERTopic framework~\cite{Bertopic}. We used the multilingual BGE-M3 model, which can embed text at different levels of granularity (sentence, passage, document) and supports over 100 languages, including those under study. News articles exceeding the 8,192-token context limit are truncated. We used dense embeddings to be compatible with BERTopic framework.

In the process of topic modeling, several parameters were systematically tuned to optimize dimensionality reduction using UMAP~\cite{McInnes2018} and subsequent dense clustering performed by HDBSCAN~\cite{HDBSCAN}. This tuning involved exploring a defined range of values for the key parameters governing these steps (see Table~\ref{tab:umap_hdbscan_params}).

\setlength{\tabcolsep}{3pt}
\begin{table}[h]
    
    \centering
    \rowcolors{2}{}{gray!15}
    {\small
    \begin{tabular}{ll}
        \toprule
        \textbf{Parameter} & \textbf{Values} \\
        \midrule
        \multicolumn{2}{c}{\textbf{UMAP Parameters}} \\
        n\_components & {[}5, 7{]} \\
        min\_dist & {[}0.05, 0.1{]} \\
        neighbors & {[}15, 30, 50{]} \\
        \midrule
        \multicolumn{2}{c}{\textbf{HDBSCAN Parameters}} \\
        min\_sample\_fraction & {[}0.3, 0.5, 0.75{]} \\
        min\_cluster\_size & {[}30, 50, 100{]} \\
        cluster\_selection\_method & {[}\enquote{leaf}, \enquote{eom}{]} \\
        \bottomrule
    \end{tabular}
    }
    \caption{Hyperparameter grid for UMAP and HDBSCAN.}
    \label{tab:umap_hdbscan_params}
\end{table}

 Then, we qualitatively identified configurations that yielded meaningful and coherent topic structures for each language split. The final set of parameters that were retained for each language split can be found in Table~\ref{tab:optimal_params}.

\setlength{\tabcolsep}{3pt}
\begin{table}[h]
    
    \centering
    \rowcolors{2}{}{gray!15}
    {\small
    \begin{tabular}{lccc}
        \toprule
        \textbf{Parameter} & \textbf{German} & \textbf{French} & \textbf{Italian} \\
        \midrule
        n\_components & 7 & 7 & 7 \\
        min\_dist & 0.1 & 0.05 & 0.1 \\
        neighbors & 15 & 30 & 30 \\
        min\_sample\_fraction & 0.3 & 0.3 & 0.3 \\
        min\_cluster\_size & 100 & 30 & 30 \\
        cluster\_selection\_method & eom & eom & eom \\
        \bottomrule
    \end{tabular}
    }
    \caption{Best hyperparameters found per language.}
    \label{tab:optimal_params}
\end{table}

Further details and any additional preprocessing steps can be provided upon request.

\section{Computation Resources}

The main compute-intensive steps in the analysis were as follows:

\begin{itemize}
    \item \textbf{NER Extraction:} Entities were extracted for 57 events, each requiring approximately 2 GPU-hours on an RTX 3090 ($\sim$0.4kWh).
    \item \textbf{Topic Modeling:} Dense embedding inference with BGE-M3 runs once, taking approximately 5 GPU-hours on A100s ($\sim$1.5kWh). Clustering with BERTopic runs on CPU. 
    \item \textbf{Entity Linking and Sentiment Analysis:} Entity-linking is time-consuming, and mGenre's guided generation hinders GPU efficiency. Inference took $\sim$96 GPU-hours for linking and $\sim$40 GPU-hours for sentiment on an RTX 2080Ti cluster. Effective energy use was 3.8kWh for linking and 8kWh for sentiment.
\end{itemize}

All computations were performed on local or institutional compute resources. Details for each step can be provided upon request.

\section{BM25 Keyword Sets and Retrieval Protocol}
\label{app:bm25}

\textbf{Protocol.} We built language-specific BM25 indexes over the CCNews-derived corpus and executed Boolean queries per language without a score cut-off. Following a precision-first strategy over recall, we designed compact, expert-curated query sets around canonical event names and widely used acronyms, enumerated salient variants, and applied minimal disambiguation where obvious ambiguity was observed. We validated relevance through qualitative spot checks of random samples. Candidates that did not meet cross-language coverage (three languages), sufficient per-language volume, or acceptable qualitative precision were dropped.

\textbf{Final multilingual queries (with a non-exhaustive list of examples not retained in the study to illustrate the strategy).}
\begin{itemize}
\item Brexit
\begin{itemize}
\item FR: \texttt{"brexit"}
\item DE: \texttt{"brexit"}
\item IT: \texttt{"brexit"}
\end{itemize}
\item Christmas
\begin{itemize}
\item FR: \texttt{"noël"}
\item DE: \texttt{"weihnachten"}
\item IT: \texttt{"natale"}
\end{itemize}
\item UN climate conferences
\begin{itemize}
\item FR: \texttt{"cop25"}; \texttt{"cop26"}; \texttt{"cop27"}; \texttt{"conférence des parties"}
\item DE: \texttt{"cop25"}; \texttt{"cop26"}; \texttt{"cop27"}; \texttt{"klimakonferenz"};
\item IT: \texttt{"cop25"}; \texttt{"cop26"}; \texttt{"cop27"}
\end{itemize}
\item Grape harvest
\begin{itemize}
\item FR: \texttt{"vendanges"}
\item DE: \texttt{"weinlese"}
\item IT: \texttt{"vendemmia"}
\end{itemize}
\item Feminist strikes
\begin{itemize}
\item FR: \texttt{"grève féministe"}
\item DE: \texttt{"frauenstreik"}
\item IT: \texttt{"sciopero femminista"}
\end{itemize}
\item 5G
\begin{itemize}
\item FR: \texttt{"5g"}
\item DE: \texttt{"5g"}
\item IT: \texttt{"5g"}
\end{itemize}
\end{itemize}

\textbf{Candidates not retained.} We considered additional candidates (e.g. WEF, WM/Weltmeisterschaft, Berlinale, Pierre Maudet affair, Crédit Suisse), but excluded them due to limited cross-language coverage (especially in Italian), insufficient per-language volume, or unstable precision under compact high-precision queries.

\section{Thematic Topic Selection (BERTopic)}
\label{app:bertopic_selection}

From the BERTopic clusters obtained per language (Appendix~B), we retained only topics that were (i) qualitatively interpretable by the authors with a meaningful coverage in the Swiss media ecosystem, (ii) of adequate size to support downstream analyses, and (iii) present in at least two language regions. We prioritized topics exhibiting a visible temporal structure over 2019–2023. Topics failing interpretability, size, or cross-language presence were dropped.

\section{Cue Lexica and Linking Robustness}
\label{app:linking_robustness}
We use compact, language-specific cue lexica to boost recall of Swiss anchoring beyond entity linking by adding a small bonus (+0.1, clipped at 1.0) to $p_{\text{swiss}}$ when a cue is detected in the article text. Patterns are implemented as case-insensitive regular expressions and comprise (i) institutional and political cues and (ii) canton names in the respective language.

French (examples):
\begin{itemize}
\item Institutional/political cues: conseil fédéral, confédération, canton, cantons, CHF, franc suisse, francs suisses, UDC, PS, PLR, Le Centre, PVL, Verts, accord bilatéral, accords bilatéraux, accord-cadre, Rahmenabkommen.
\item Cantons: Zurich, Berne, Lucerne, Uri, Schwyz, Obwald, Nidwald, Glaris, Zoug, Fribourg, Soleure, Bâle-Ville, Bâle-Campagne, Schaffhouse, Appenzell Rhodes-Extérieures, Appenzell Rhodes-Intérieures, Saint-Gall, Grisons, Argovie, Thurgovie, Tessin, Vaud, Valais, Neuchâtel, Genève, Jura.
\end{itemize}

German (examples):
\begin{itemize}
\item Institutional/political cues: Bundesrat, Eidgenossenschaft, Kanton, Kantone, CHF, Schweizer Franken, SVP, SP, FDP, Die Mitte, Gr\"une, GLP, bilaterales Abkommen, bilaterale Abkommen, Rahmenabkommen.
\item Cantons: Z\"urich, Bern, Luzern, Uri, Schwyz, Obwalden, Nidwalden, Glarus, Zug, Freiburg, Solothurn, Basel-Stadt, Basel-Landschaft, Schaffhausen, Appenzell Ausserrhoden, Appenzell Innerrhoden, St. Gallen, Graub\"unden, Aargau, Thurgau, Tessin, Waadt, Wallis, Neuenburg, Genf, Jura.
\end{itemize}

Italian (examples):
\begin{itemize}
\item Institutional/political cues: Consiglio federale, Confederazione, cantone, cantoni, CHF, franco svizzero, franchi svizzeri, UDC, PS, PLR, Il Centro, Verdi, PVL, accordi bilaterali, accordo quadro, Rahmenabkommen.
\item Cantons: Zurigo, Berna, Lucerna, Uri, Svitto, Obvaldo, Nidvaldo, Glarona, Zugo, Friburgo, Soletta, Basilea Città, Basilea Campagna, Sciaffusa, Appenzello Esterno, Appenzello Interno, San Gallo, Grigioni, Argovia, Turgovia, Ticino, Vaud, Vallese, Neuchâtel, Ginevra, Giura.
\end{itemize}

\noindent Linking-threshold robustness. We recomputed domestication profiles and LPSR with mGenre confidence thresholds at 0.6, 0.7, and 0.8 (baseline 0.5). Aggregate series and qualitative conclusions were unchanged across thresholds.

\section{Relative-Change Robustness for Lexical Metrics}
\label{app:lexical_relchange}
We compute relative change as percent deviation from the language-specific baseline mean for each metric $m \in \{\text{TTR, LD, MSL}\}$:
\begin{equation*}
r_m(t) \;=\; 100 \times \frac{x_m(t) - \mu_m}{\mu_m}\,,
\end{equation*}
and then average monthly (three-month rolling mean). The temporal ordering and cross-language contrasts mirror those obtained with standardized $z$-scores, magnitudes differ as expected.

\begin{figure}[h]
  \centering
  \includegraphics[width=0.95\linewidth]{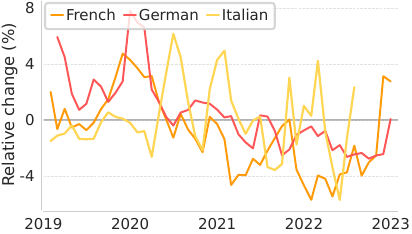}
  \caption{Brexit: relative-change composite lexical index by language (monthly, three-month rolling mean).}
\end{figure}

\begin{figure}[h]
  \centering
  \includegraphics[width=0.95\linewidth]{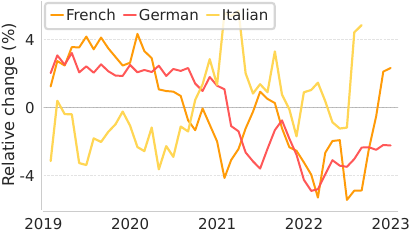}
  \caption{Swiss Wolf: relative-change composite lexical index by language.}
\end{figure}

\begin{figure}[h]
  \centering
  \includegraphics[width=0.95\linewidth]{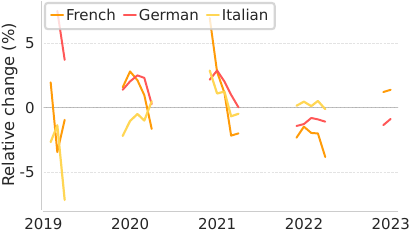}
  \caption{Christmas (Nov–Mar): relative-change composite lexical index by language.}
\end{figure}

\begin{figure}[h]
  \centering
  \includegraphics[width=0.95\linewidth]{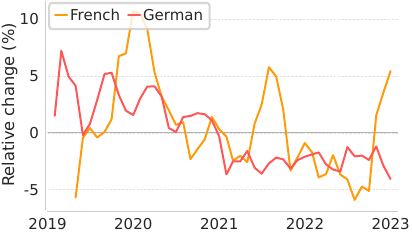}
  \caption{British Royal Family: relative-change composite lexical index by language.}
\end{figure}

\section{GLiNER Benchmarks and Cross-language Comparability}\label{app:gliner_bench}

The GLiNER model has been extensively evaluated across benchmarks (Zaratiana et al., 2023). On the out-of-domain NER benchmark (seven datasets), the large variant achieves an average F1 of 60.9, outperforming instruction-tuned LLMs such as UniNER-13B (55.6) and InstructUIE-11B (47.2) while using far fewer parameters (Table~1 in Zaratiana et al., 2023). Across 20 additional NER datasets, GLiNER-L attains a 47.8 average zero-shot F1 and 58.9 on the multilingual WikiANN dataset (Table~2). For multilingual robustness on complex settings, GLiNER-Multi (multilingual backbone) reports competitive zero-shot results on MultiCoNER for high-resource Latin-script languages, with F1 of 39.5 (German) and 42.1 (Spanish), exceeding ChatGPT on most languages in that benchmark (Table~3). To further mitigate any residual model or language effects in our study, we: (i) mainly use comparisons to canonical entity types (PER/ORG/LOC), (ii) perform Wikidata linking and retain only linked mentions (mGenre conf.\ $\geq$ 0.5), and (iii) aggregate at weekly/monthly levels and use proportion- or ratio-based indicators (e.g. domestication profiles, LPSR), emphasizing within-language contrasts over absolute levels. 

\section{Plotting Conventions and CPD Implementation Details}\label{app:plots}
For display, we plot weekly counts with a three-week moving average and a LOESS trend. Some events are not consistently covered across all language subsets over the full window (most notably in Italian). To avoid spurious fluctuations, we truncate a language series when the per-period sample is too small to be reliable, affected figures therefore show gaps or early terminations. Captions flag restricted windows where applicable (e.g. Christmas lexical plots show only November–March).
\smallskip

\noindent Change-point detection (implementation). CPD is run on the three-week–smoothed series. We generate candidate change points using multiple detectors, PELT with an RBF cost across a small penalty grid, Binary Segmentation with the number of breakpoints matched to PELT’s baseline, and Bayesian Online CPD with a constant hazard. We cluster all candidate cut points within ±1 week and retain the cluster median as a consensus change point when supported by at least two detectors. In figures, consensus CPs are shown as shaded ±1 week bands.

\section{Dossier: Christmas}
\label{app:dossier_christmas}

\begin{figure}[h]
    \centering
    \includegraphics[width=0.45\textwidth]{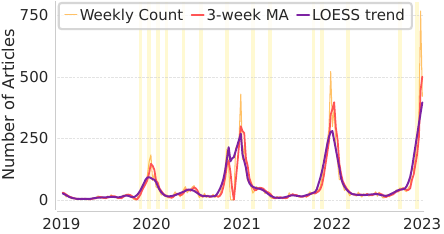}
    \caption{Weekly distribution of \textit{Christmas}-related news articles across all languages, with change points detected (consensus bands). For Christmas, lexical plots display only Nov–Mar when the sample is meaningful.}
    \label{fig:app_christmas_article_distribution}
\end{figure}

\paragraph{Articles over time and change points.} Figure~\ref{fig:app_christmas_article_distribution} shows the expected calendar‑driven pattern: attention rises in late November, peaks in December, and drops sharply in early January. The consensus CPD bands consistently bracket the beginning of Advent and the post‑holiday decline, indicating stable seasonal segmentation across years. Peak amplitude grows over the period, with particularly pronounced spikes in 2021–2022, this coincides with renewed public interest and logistics/regulation angles (e.g. reopening of markets, pandemic‑related rules), but may also reflect a gradual shift toward more digital service coverage.

\begin{figure}[h]
    \centering
    \includegraphics[width=0.45\textwidth]{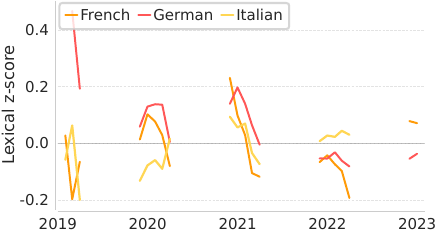}
    \caption{Evolution of the standardized composite lexical index for \textit{Christmas} in French, German, and Italian Swiss media (monthly averages, three-month rolling mean).}
    \label{fig:app_christmas_lexical_metrics}
\end{figure}

\paragraph{Standardized lexical complexity.} The composite standardized lexical index in Figure~\ref{fig:app_christmas_lexical_metrics} varies only modestly around zero, consistent with a recurrent event covered largely through service pieces, event guides, and features. Slight positive departures in late 2019–2020 (notably in German and French) align with explanatory items on restrictions, supply chains, or public‑health guidance, whereas 2021–2022 shows a mild drift toward simpler, more practical formats (lists, market guides). Overall, differences are small, suggesting stability in register with episodic excursions when contextual or policy information becomes salient.

\begin{figure}[h]
    \centering
    \includegraphics[width=0.99\linewidth]{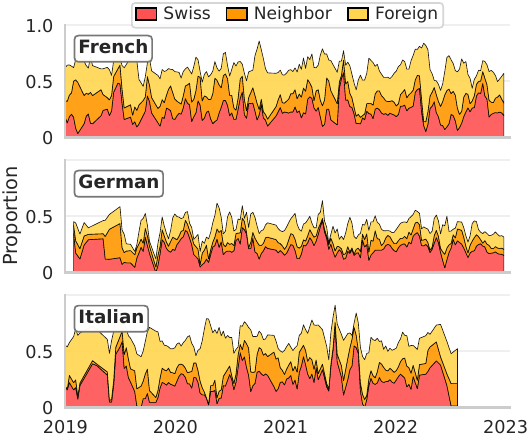}
    \caption{Weekly domestication profile for \textit{Christmas} by language.}
    \label{fig:app_christmas_domestication}
\end{figure}

\begin{figure}[h]
    \centering
    \includegraphics[width=0.99\linewidth]{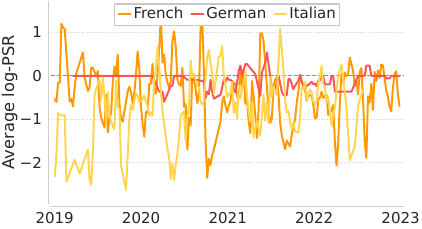}
    \caption{LPSR over time for \textit{Christmas} by language (weekly averages).}
    \label{fig:app_christmas_lpsr}
\end{figure}

\paragraph{Domestication profile and LPSR.} Combining the domestication stacks and LPSR (Figures~\ref{fig:app_christmas_domestication} and~\ref{fig:app_christmas_lpsr}) shows that the topic is primarily domesticated at the peak: Swiss anchoring rises around December as outlets emphasize local markets, municipal events, and retail. A persistent \enquote{foreign‑other} layer reflects global retail and travel references, whereas the neighbor share is comparatively small. The weekly log‑PSR is mostly negative in all languages, indicating that when foreign actors appear they are more often non‑proximate than proximate, short‑lived upward bumps occur around cross‑border markets or policy comparisons, with slightly more frequent lifts in the French series. This pattern is consistent with a local service agenda supplemented by global, not neighbor‑centric, cues.


\begin{figure}[h]
    \centering
    \includegraphics[width=0.99\linewidth]{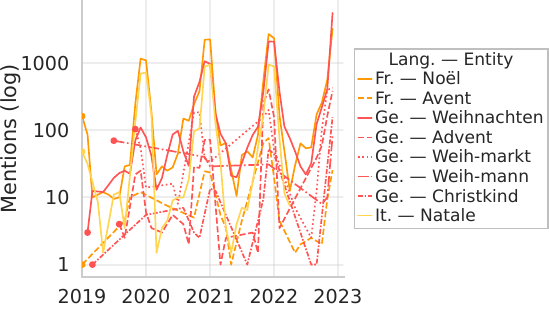}
    \caption{Monthly frequency of named entities related to \textit{Christmas} across French, German, and Italian Swiss media.}
    \label{fig:app_christmas_ner_timeline}
\end{figure}

\begin{figure}[h]
    \centering
    \includegraphics[width=0.99\linewidth]{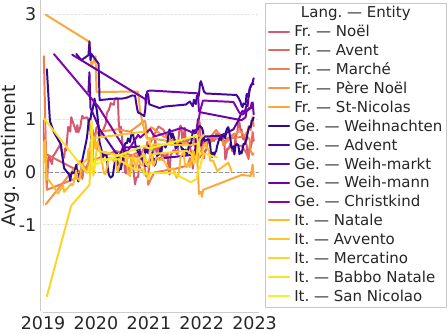}
    \caption{Evolution of monthly sentiment scores for key entities in \textit{Christmas} coverage (by language).}
    \label{fig:app_christmas_ner_sentiment_timeline}
\end{figure}

\paragraph{Key entities and sentiment.} Entity timelines (Figure~\ref{fig:app_christmas_ner_timeline}, log scale) follow the Advent–Christmas cycle, with richer lexical differentiation in German (\emph{Weihnachtsmarkt}, \emph{Advent}, \emph{Christkind}, \emph{Weihnachtsmann}) and smoother profiles for French (\emph{Noël}, \emph{Avent}) and Italian (\emph{Natale}). Targeted sentiment (Figure~\ref{fig:app_christmas_ner_sentiment_timeline}) is predominantly neutral to mildly positive, reflecting lifestyle and culture frames. Temporary dips in 2020–2021 for market‑related terms align with cancellations and restrictions, sentiment rebounds thereafter, turning clearly positive for several German entities by late 2022, while French and Italian remain closer to neutral with small positive drifts.

\paragraph{Brief narrative.}  Taken together, the Christmas dossier depicts a recurrent, institutionally timed attention cycle: coverage builds with Advent, peaks in December, and recedes immediately after the holidays. Language use is stable and service‑oriented, with only brief complexity upticks when contextual or regulatory information becomes newsworthy (notably in 2020–2021). Domestication intensifies at the peak as outlets foreground local markets and municipal events, while LPSR remains mostly negative, foreign references are present but global rather than neighbor‑centric, with only episodic proximate lifts. Entity salience reflects the same seasonal rhythm and a richer lexical differentiation in German, and targeted sentiment stays neutral to mildly positive, dipping during cancellations and rebounding as in‑person festivities return. Overall, Christmas functions as a strongly domesticated, culturally shared event across language regions, with cross‑linguistic variation expressed more in vocabulary than in framing.

\section{Dossier: British Royal Family}
\label{app:dossier_brf}

The British Royal Family emerged as a high-quality thematic topic in French and German, but did not reach representation thresholds in Italian. We therefore report it as a two-language case.

\begin{figure}[h]
    \centering
    \includegraphics[width=0.45\textwidth]{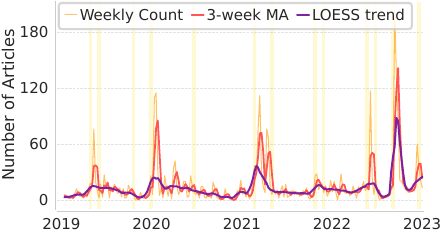}
    \caption{Weekly distribution of \textit{British Royal Family}-related news articles across French and German, with change points detected (consensus bands).}
    \label{fig:app_british_article_distribution}
\end{figure}

\paragraph{Articles over time and change points.} Figure~\ref{fig:app_british_article_distribution} exhibits a sequence of sharp, event‑driven surges. Consensus bands cluster around well‑known milestones: the couple’s step‑back announcement (January2020), the death and funeral of Prince Philip (April2021), the Platinum Jubilee (June2022), and the death of Queen ElizabethII with the accession of CharlesIII (September2022, the global maximum. A further rise appears in December~2022 with streaming releases and related coverage. Between these spikes, the LOESS trend returns to a low steady baseline, indicating a celebrity/ceremonial agenda punctuated by high‑visibility moments rather than a sustained beat.

\paragraph{Standardized lexical complexity.} The lexical index varies only modestly around zero in both French and German coverage. French shows brief above-baseline bursts (early 2020 - mid 2022) that quickly revert to baseline or below, consistent with waves of explanatory or commemorative pieces around high-visibility moments. German starts slightly higher in early 2019 but remains mostly below baseline from 2021 onward, suggesting a predominance of short updates outside peak phases.

\begin{figure}[h]
    \centering
    \includegraphics[width=0.45\textwidth]{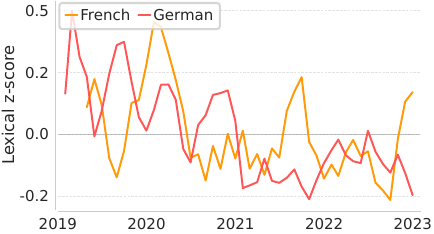}
    \caption{Evolution of the standardized composite lexical index for \textit{British Royal Family} in French and German Swiss media (monthly averages, three-month rolling mean).}
    \label{fig:app_british_lexical_metrics}
\end{figure}

\begin{figure}[h]
    \centering
    \includegraphics[width=0.45\textwidth]{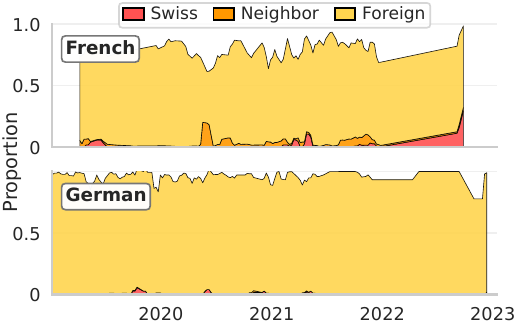}
    \caption{Weekly domestication profile for \textit{British Royal Family} by language.}
    \label{fig:app_british_domestication}
\end{figure}

\begin{figure}[h]
    \centering
    \includegraphics[width=0.45\textwidth]{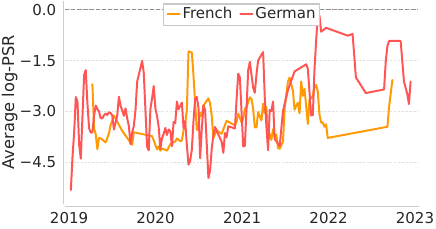}
    \caption{LPSR over time for \textit{British Royal Family} by language (weekly averages).}
    \label{fig:app_british_lpsr}
\end{figure}

\paragraph{Domestication profile and LPSR.} The domestication stacks and LPSR (Figures~\ref{fig:app_british_domestication} and~\ref{fig:app_british_lpsr}) point to a predominantly foreign frame in both languages. \enquote{Foreign‑other} (which includes the UK and supranational entities) saturates coverage, Swiss anchoring is minimal and neighbor shares are marginal. Consistently, weekly log‑PSR values are strongly negative throughout, indicating that foreign references, when present, are overwhelmingly non‑proximate rather than tied to France or Germany/Austria. Occasional, short‑lived upticks reflect cross‑border angles (e.g. travel, media distribution), but they do not alter the overall international celebrity framing.


\begin{figure}[h]
    \centering
    \includegraphics[width=0.45\textwidth]{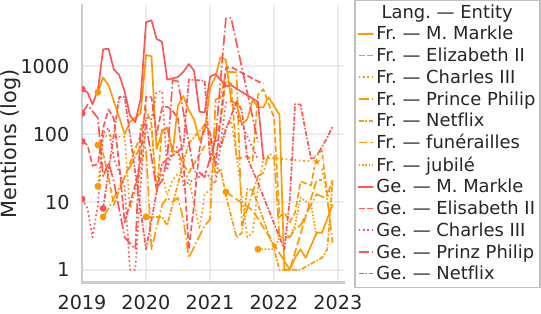}
    \caption{Monthly frequency of named entities in French and German coverage of the \textit{British Royal Family}. The star marks the first mention of \enquote{Charles III} in the corpus.}
    \label{fig:app_british_ner_timeline}
\end{figure}

\begin{figure}[h]
    \centering
    \includegraphics[width=0.45\textwidth]{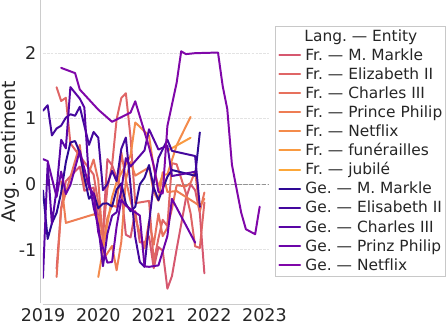}
    \caption{Evolution of monthly sentiment scores for key entities in \textit{British Royal Family} coverage (by language).}
    \label{fig:app_british_ner_sentiment_timeline}
\end{figure}

\paragraph{Key entities and sentiment.} Entity timelines in Figure~\ref{fig:app_british_ner_timeline} align closely with the change‑points. Mentions of Meghan Markle spike around January~2020, Prince Philip surges in April~2021 alongside \enquote{funérailles}, Queen Elizabeth~II dominates in June~2022 (Jubilee) and reaches the series’ apex in September~2022, \enquote{Charles~III} emerges from that point onward. References to \enquote{Netflix} climb near major releases (e.g. \enquote{The Crown} seasons, late‑2022 documentary), especially in French coverage. Targeted sentiment (Figure~\ref{fig:app_british_ner_sentiment_timeline}) is heterogeneous: the Queen and ceremonial labels trend neutral‑to‑positive (commemorative tone), Meghan Markle is more negative on average in both languages during controversy windows, and \enquote{Netflix} skews mildly positive, particularly in German, around entertainment features, with dips when coverage turns to criticism or controversy. These trajectories reinforce a media logic centered on major ceremonies and media tie‑ins rather than sustained policy or institutional angles.

\paragraph{Brief narrative.}
Overall, the British Royal Family dossier portrays an internationally framed, personality‑driven topic punctuated by ceremonial and media events. Attention concentrates around a small set of global milestones, domestication remains negligible, and cultural‑proximity cues are weak. Entity salience and sentiment track these bursts, respectful around commemorations, more polarized around celebrity controversies, and mildly positive around streaming releases, underscoring a transnational, entertainment‑adjacent news logic in both French‑ and German‑language Swiss outlets.

\end{document}